\documentclass[letterpaper]{article} 
\usepackage{aaai25}  
\usepackage{times}  
\usepackage{helvet}  
\usepackage{courier}  
\usepackage[hyphens]{url}  
\usepackage{graphicx} 
\usepackage{natbib}  
\usepackage{caption} 
\usepackage{algorithm}
\usepackage{algorithmic}
\usepackage{amsmath}
\usepackage{amsfonts}
\usepackage{multirow}
\usepackage{graphicx}
\usepackage{multirow}
\usepackage{rotating}
\usepackage{subcaption}
\usepackage{graphicx}
\usepackage{subcaption}

\usepackage{newfloat}
\usepackage{listings}
\DeclareCaptionStyle{ruled}{labelfont=normalfont,labelsep=colon,strut=off} 
\floatstyle{ruled}
\newfloat{listing}{tb}{lst}{}
\floatname{listing}{Listing}
\title{Membership Inference Test: \\
Auditing Training Data in Object Classification Models}
\author{
    Gonzalo Mancera,
    Daniel DeAlcala,
    Aythami Morales,
    Ruben Tolosana,
    Julian Fierrez
}
\affiliations{
    \textsuperscript{\rm 1}Biometrics and Data Pattern Analytics Lab, Universidad Autonoma de Madrid, Spain\\

    \email{\{gonzalo.mancera, daniel.dealcala,aythami.morales,\\\ ruben.tolosana, julian.fierrez\}@uam.es}
}
\newcommand{\email}[1]{\texttt{#1}}

\usepackage{bibentry}

\begin{document}

\maketitle

\begin{abstract} \label{sec:abstract}
In this research, we analyze the performance of Membership Inference Tests (MINT), focusing on determining whether given data were utilized during the training phase, specifically in the domain of object recognition. Within the area of object recognition, we propose and develop architectures tailored for MINT models. These architectures aim to optimize performance and efficiency in data utilization, offering a tailored solution to tackle the complexities inherent in the object recognition domain. We conducted experiments involving an object detection model, an embedding extractor, and a MINT module. These experiments were performed in three public databases, totaling over 174K images. The proposed architecture leverages convolutional layers to capture and model the activation patterns present in the data during the training process. Through our analysis, we are able to identify given data used for testing and training, achieving precision rates ranging between 70\% and 80\%, contingent upon the depth of the detection module layer chosen for input to the MINT module. Additionally, our studies entail an analysis of the factors influencing the MINT Module, delving into the contributing elements behind more transparent training processes.

\end{abstract}

\section{Introduction} \label{sec:intro}

In recent years, Artificial Intelligence (AI) has experienced rapid deployment in virtually all aspects of our society. This expansion has been driven by significant advances in machine learning and large-scale data processing. However, this rapid progress has raised ethical and legal challenges that require urgent attention. In response to these pressing concerns, the European Union has taken proactive steps by introducing new legislation in March 2024\footnote{\url{https://artificialintelligenceact.eu}}, specifically targeting the regulation of AI usage within the region. The proposed legislation underscores a unified effort to address the ethical and societal repercussions associated with the expanding prevalence of AI/ML technologies. It is crafted with the overarching aim of protecting citizens' fundamental rights through the implementation of measures aimed at fostering transparency and accountability throughout the development and implementation stages of AI/ML systems.

Consequently, new auditing instruments need to be designed to oversee AI technologies and their proper integration into our society. Understanding how these systems are trained is crucial for evaluating their performance and implications. In particular, the AI training process involves using large datasets to teach algorithms to recognize patterns and make decisions. Therefore, it is essential to develop auditing tools that allow for a rigorous examination of training datasets and protocols. Membership Inference Attacks (MIAs)\cite{shokri2017membership} involve attackers trying to access confidential information about the data used to train a model, such as sensitive medical data. The literature demonstrates the feasibility of such attacks and proposes solutions to prevent them. A new approach, called Membership Inference Test (MINT) proprosed in \cite{dealcala2024my}, focuses on detecting unauthorized data usage in models, aiming to identify if given data was used without users consent. MINT audits AI/ML models to improve transparency and explainability for users \cite{BARREDOARRIETA202082,2023_ECAI}, allowing individuals to determine whether specific data (such as personal \cite{2017_Access_HEmultiDTW_Marta} or some other sensitive data \cite{2021_TPAMI_SensitiveNets_Morales}) contributed to the development of these models. Both MIAs and MINT address the task of membership inference, but operate in different contexts.

Once a model has been trained, it is not possible to assess the data used during the training process. There is a need to develop new tools capable of determining if given data was used to train AI models. In this context, the primary contributions can be outlined as follows: 

 \begin{itemize}
    \item We analyze the performance of Membership Inference Test (MINT) for objects classification models based on Convolutional Neural Networks (CNNs). The MINT models were trained to determine if specific images were used during the training process of these object detection models. 
    \item We provide an in-depth evaluation of the MINT model in the object classification domain. The experiments were conducted using a total of three public datasets, comprising a combined total of 174K images. Additionally, various experimental scenarios were considered. The results achieved a performance of around 80\% accuracy, highlighting different factors that affect the detection of test and train samples.
\end{itemize}

The remainder of the paper is structured as follows: Section \ref{sec:related} offers a review of state-of-the-art studies relevant to the paper's scope. Section \ref{sec:mintaproach} delves into the description of the proposed approach and experimental framework Section \ref{sec:framework} explains the experimental protocol employed and databases. Results are analyzed and discussed in Section \ref{sec:experiments}, while conclusions and avenues for future research are presented in Section \ref{sec:conclusion}.

\section{Related Works} \label{sec:related}
\begin{figure}[t!]

\centering
\includegraphics[width=0.49\textwidth]{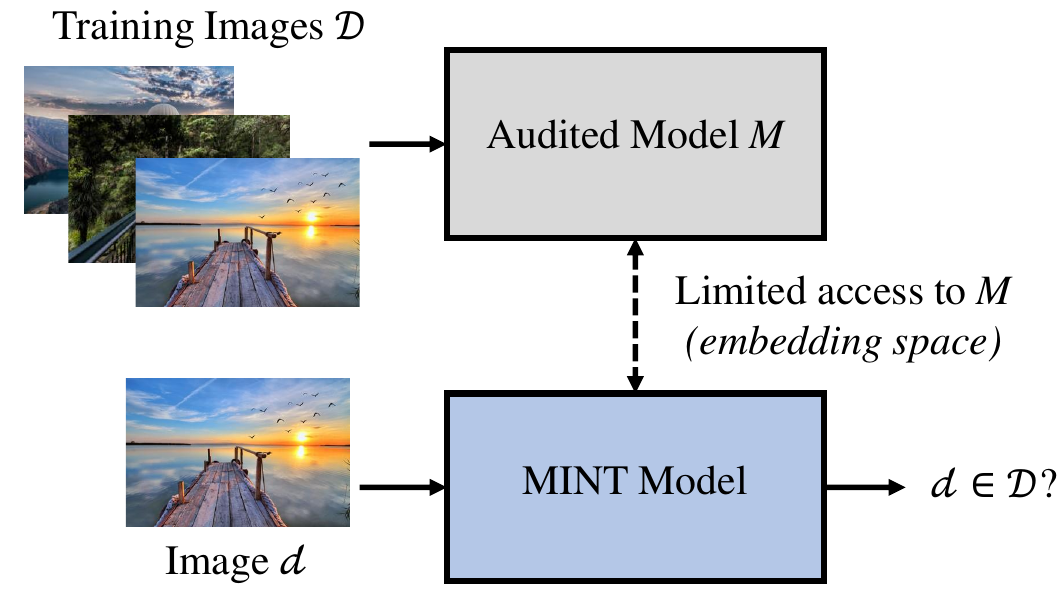}
\caption{The objective of MINT is to discern whether a given data point (\textit{d}) was utilized in the training process of an audited AI Model (\textit{M}) trained with a specific database ($\mathcal{D}$).}
\label{fig:esquemamint}
\end{figure}

Our work falls under the \textit{``Membership Inference Test (MINT)''} framework. This approach is rooted in the \textit{``Membership Inference Attacks (MIAs)''} research stream, which aims to exploit information leakage from models, particularly regarding their training data \cite{shokri2017membership}. By aligning with MIAs principles, we aim to understand models susceptibility to exploit training traces and developing auditing tools in the object classification domain (see Fig.\ref{fig:esquemamint}).

\subsection{Membership Inference Attacks}
Comprehending the data employed for training AI models raises significant security concerns, potentially exposing confidential or private details spanning various domains like consumer inclinations \cite{zhang2021graph} or medical archives \cite{chen2020differential,zhang2022membership} among others. The fundamental principle underlining the research presented herein is the extraction of sensitive information \cite{2021_TPAMI_SensitiveNets_Morales} through an adversarial methodology, even when devoid of access to the model or its training data, due to the providers' hesitantness to disclose such information. The pioneering investigation by Shokri et al. \cite{shokri2017membership} laid the foundation for this research trajectory. In their methodology, they employ shadow models to mimic the functionalities of the original model, despite not having direct access to it. This indicates that instead of directly working with the original model, researchers use secondary models constructed and trained to replicate its behavior \cite{hu2022membership}.
This strategy provides full authority over both the training and nontraining samples of the dataset utilized by these models. A binary classifier is trained to distinguish between the data employed for training and the test data in the shadow models. Images were processed through the network, and resulting embeddings from shadow models were employed to train a binary classifier.

 Nasr et al. \cite{nasr2019comprehensive} in  their research introduced the Black-box and White-box terminology into this domain. Previous endeavors predominantly relied on the output of shadow models for classification, essentially the output embedding (Black-box). However, Nasr et al. proposed a White-box framework, affording them access to the activations, loss, and gradients. The researchers illustrated that access to White-box information provides limited usefulness in discerning whether a sample was utilized for training or not. The most favorable outcomes were attained employing gradients \cite{nasr2019comprehensive}. Generally, within the White-box framework, there is a scarcity of studies in the literature, and no significant findings have emerged.

\subsection{Membership Inference Test}
MINT serves as a Membership Inference Test designed to scrutinize the adherence of neural networks to prevailing regulatory standards governing their training data. This process necessitates a certain degree of cooperation from the model owner to ensure compliance with legal frameworks. The overarching objective is to devise methodologies applicable across diverse data modalities (text, images, audio, etc.), accommodating different amounts of information provided by the model developer about the model training and operation.. These range from scenarios where minimal or no data information is available up to full detailed information \cite{dealcala2024my}. 

In the context of MINT, we take on the responsibility of model auditors, having access to either the original model or limited insights into its structure. This aligns with the practical aspect of our research, especially our efforts to adapt to evolving legal regulations and create advanced data protection solutions. Importantly, this method diverges from MIAs, as they avoid the need to create ``shadow models'' to mimic the original model's actions. Instead, by having access to the original model, we can apply MINT techniques directly, which may lead to varying detection results for comparable tasks.

\section{Membership Inference Test (MINT): Proposed Approach} \label{sec:mintaproach}

In the MINT framework, our role as model auditors involves scrutinizing the inner workings of the original model, or at least having access to some extent of information about it. This is a crucial aspect, particularly within the context of our work's focus on adapting to emerging legal frameworks and crafting innovative data protection tools. Unlike traditional approaches like MIAs, where the need arises to create shadow models to replicate the behavior of the original model, MINT takes a different path. By having direct access to the original model, we bypass the complexities and uncertainties associated with shadow modeling.
\begin{figure*}[h]
\centering
\includegraphics[width=\textwidth]{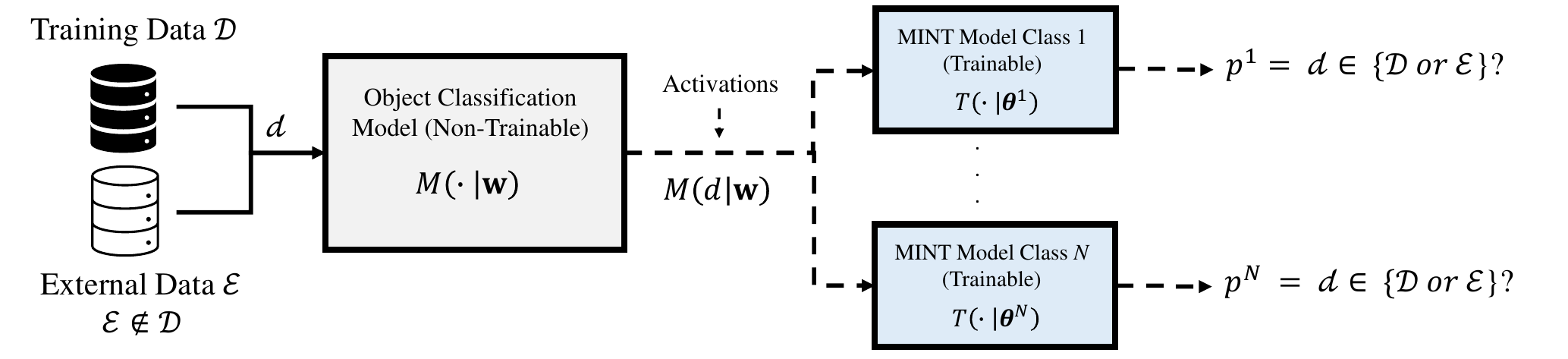}
\caption{The Membership Inference Test (MINT) Model ($\mathcal{T}$) is trained to ascertain whether a given data instance ($d$) was included in the training dataset $\mathcal{D}$ utilized by an Audited Artificial Intelligence Model (\textit{M}). The inputs for the MINT Model comprise Auxiliary Auditable Data, such as activation maps corresponding to data samples $d$, as well as the predictions made by model \textit{M}.}
\label{fig:esquemamint_ext}
\end{figure*}

This distinction allows us to directly apply our techniques and methodologies to the original model, leading to potentially different outcomes in terms of detection and analysis, even when addressing similar tasks. With this direct access, we gain deeper insights into the model's behavior, vulnerabilities, and biases, enabling us to refine our approaches more effectively and tailor them to the specific nuances of the original model. Ultimately, this approach not only streamlines the auditing process, but also enhances the accuracy and relevance of our findings in addressing evolving challenges in data protection and compliance with regulatory standards.

\subsection{Problem Statement}

Let us consider the Training Data ($\mathcal{D}$), an External Data $\mathcal{E}$, and a compilation of samples $d$ $\in$ $\mathcal{D}$ $\cup$ $\mathcal{E}$. We also consider a trained model $M$ train for a particular task using the dataset $\mathcal{D}$. For each input data instance $d$, the model $M$ produces a result $y$ using $d$ and a collection of parameters $\textbf{w}$ acquired during the training phase, expressed as $M$($d$$\mid$$\textbf{w}$) (see Fig.\ref{fig:esquemamint_ext}).

We propose that an authorized auditor has access to the model $M$, allowing him to gather information on how $M$ processes data $d$. This insight includes the resulting outcome $y$ = $M$($d$$\mid$$\textbf{w}$). The outcome doesn't necessarily have to be the final layer of the object classifier model, as will be illustrated later in Section \ref{sec:experiments}. It could even be any intermediate layer.

The goal of the Membership Inference Test is to ascertain whether a given data instance $d$ was part of the training data $\mathcal{D}$ used to train model $M$. To achieve this, an authorized entity utilizes Supplementary Verifiable Data and/or the outcome $y$ to train a MINT auditing model $T$($\cdot$$\mid$$\theta^{i}$), where $i \in \{1,...,N\}$ denotes the object class (i.e., classes available in the the object classifier). This auditing model is designed to predict whether a given data sample \textit{d} belongs to the training data $\mathcal{D}$ or the External Data $\mathcal{E}$ $\notin$ $\mathcal{D}$. This process is repeated for each object class. MINT models leverage the memorization capabilities inherent in machine learning processes. The fundamental elements of MINT are as follows:
\begin{itemize}
    \item Audited Model $M$: a trained model defined by an architecture and a set of parameters \textbf{$w$}.
    \item External Data $\mathcal{E}$: any data out of the collection ($\mathcal{D}$).
    \item Training Data $\mathcal{D}$: collection of data used to train $M$.
    \item Model Prediction $y$ = $M$($d$$\mid$$\textbf{w}$): final outcome of \textit{M} that
    results from processing an input data $d$ using the set of the parameters $\textbf{w}$, depends on the layer as output on the Object Classifier.
    \item MINT Model $T$: as many models as classes defined by an architecture and
    a set of parameters $\theta$ trained using activations from the audited model. In this work we train a specific MINT model ($T(\cdot\mid\theta^{i})$) for each class $i$ in the audited object classification model. 
\end{itemize}

\subsection{MINT for Object Classification Domain}\label{sec:architecture}
The authors in the original work \cite{dealcala2024my} conducted experiments with an Face Recognition Model; in this case, we will examine another type of model, specifically an object classifier. Despite both being based on a CNN, these models exhibit several substantial differences: while the FR Model returns an embedding in a representation space, the object classifier outputs a classification vector. Consequently, their training methods and the type of information they output are markedly dissimilar. The authors in \cite{dealcala2024my} exclusively worked with a single MINT Model, as their study focused solely on facial images with similar characteristics. In contrast, the images in our dataset vary significantly, encompassing 10 diverse classes ranging from horses to airplanes. Thus, it is necessary to train a distinct MINT Model for each specific class. This approach ensures that each MINT Model analyzes a homogeneous class, enabling it to focus on detecting differences in activations for data used or not used during training. The workflow employed in this research comprises three distinct modules:

\begin{itemize}
    \item Object Classifier $M$: it is the model of which we want to know if given data $d$ has been used for training or not, is responsible for processing input data, in the present case, object images, and categorizing the objects present in them into different predefined classes depending on the database used. This classifier is essential for the specific task addressed in the research work, as it provides the foundation upon which membership inferences and other analytical operations will be conducted. The architecture of this Convolutional Neural Network (CNN) comprises six convolutional layers. Three sets of convolutional layers are implemented, each followed by a batch normalization layer and ReLU activation. Each set consists of two convolutional layers with a total of 32, 64, and 128 filters respectively, all with the same kernel size, $3 \times 3$. After each set of convolutional layers, a max-pooling layer is applied to reduce the output dimensionality. Additionally, dropout layers are integrated after each max-pooling layer to prevent overfitting. Subsequently, the data is flattened and connected to a dense layer with 128 units and ReLU activation. A dropout layer is then added, followed by a dense output layer with as many units as classes are in the dataset and softmax activation, suitable for multiclass classification. It is noted that the same architecture is consistently utilized across all experiments in the classifier. However, to illustrate that it works with other architectures, the subsection \ref{architeturevariation} presents a range of experiments by varying the object classifier.
    
    \item Embeddings Selector: this module allows us to select which activation layer we want to select from the model as output, which will later be used as the input layer in the MINT module.
    
    \item MINT Module: receives as input the vector representations from the Embeddings Selector $y$, and is responsible for classifying them as to whether they have been used to train the object classifier or not ($p^{i}$). This architecture consists of an input layer which size depend on the embedding size, and ReLU activation, followed by a Max Pooling layer. After the Max Pooling layer, another layer with 64 filters of the same dimensions and activation is added. The architecture is then completed with a dropout rate of 0.5, followed by the output layer with 1 neuron and sigmoid activation.
\end{itemize}

\section{Experimental Framework} \label{sec:framework}
Firstly, we present the object databases used in this study and then we will present our experimental protocol.
\subsection{Databases}\label{sec:databases}
The experiments have been conducted with the following datasets:
\begin{itemize}
    \item  \textbf{CIFAR-10} \cite{krizhevsky2009learning}. The CIFAR-10 dataset serves as a standard reference for classification tasks, is composed of 60,000 color images, each measuring $32 \times 32$ pixels, and is divided into 10 distinct categories. These categories encompass various objects such as airplanes, automobiles, birds, cats, deer, dogs, frogs, horses, ships, and trucks. This has been the database around which the largest number of experiments have revolved.
    
    \item \textbf{CIFAR-100} \cite{krizhevsky2009learning}. Alike CIFAR-10, the CIFAR-100 dataset consists of 60,000 color images, each measuring $32 \times 32$ pixels and categorized into 100 different classes, with 600 images per class. This extensive dataset covers a diverse range of objects and categories, providing a rich variety for experimentation and research.

    \item \textbf{GTSRB} (German Traffic Sign Recognition Benchmark) \cite{stallkamp2011german}. The GTSRB dataset consists of a total of 51,839 images, each showcasing 43 types of traffic signs commonly encountered on roadways in Germany. GTSRB database is designed to represent real-world scenarios as accurately as possible, which includes varying light conditions and rich backgrounds. Given that the size of each image varies, we resize them to dimensions of $32 \times 32 \times 3$. 
\end{itemize}
\subsection{Experimental Protocol}\label{sec:experimentalprotocol}

In order to devise an appropriate experimental protocol, it is crucial to consider different approaches for both the Training Data and the External Data which has not been utilized for training the Audited Model. Below, we discuss the various elements taken into account in this research to establish a fair experimental protocol:

\begin{enumerate}
    \item  As stated in Section \ref{sec:databases}, experiments are conducted on the three mentioned databases separately. Despite the images sharing the same nature within each class, intra-class images are entirely different, ensuring that the nature of the results is not compromised and the images used for train are totally different from the ones used for test.
    \item The experiments conducted are always performed using the same database; there is no mixing of databases at any point or data augmentation \cite{kaya2021does}. Both the Training Data $\mathcal{D}$ and the External Data $\mathcal{E}$ originate from the same database. Some samples are used to train the Audited Model $\mathcal{D}$, while others are excluded from the training process $\mathcal{E}$.
    \item One of the primary objectives emphasized in this work is identifying the factors that facilitate information leaks. Therefore, all experiments are conducted using the same Audited Model $M$, which is described in greater detail in Section \ref{sec:architecture} for enhanced precision.
\end{enumerate}

\subsubsection{Object Classifier Training}

The object classifier presented in Section \ref{sec:architecture} undergoes training, where the number of epochs varies depending on the experiment we are facing, ranging from 150 epochs to 5000 as reflected in Section \ref{sec:experiments}. The batch size used has always been 32; no exploration has been done to vary this parameter. The train-test split for all datasets has been 64\% of samples dedicated to train $\mathcal{D}$ and 36\% for test $\mathcal{E}$, ensuring balanced training where each of the 3 databases has balanced number of classes.

During the training process, the model's performance is primarily evaluated using accuracy as the metric. Accuracy measures the proportion of correctly predicted instances over the total number of instances in the dataset. It provides a comprehensive assessment of the model's overall correctness in its predictions. Table \ref{tab:performance} presents the classification accuracy obtained for the different databases.

\begin{table}[t!]
\caption{Classification accuracy of the object classifiers performance trained for the experimental framework.}
    \centering
   \setlength{\tabcolsep}{2,5pt}
\begin{tabular}{l|cc}
\textit{Dataset} & \textit{\begin{tabular}[c]{@{}l@{}}Training\\ Accuracy\end{tabular}} & \textit{\begin{tabular}[c]{@{}l@{}}Testing\\ Accuracy\end{tabular}} \\ \hline

CIFAR-10         & 0.992                                                               & 0.872                                                             \\
                                                         
CIFAR-100        & 0.812                                                                   & 0.552 \\
GTSRB & 0.997 & 0.991 \\ \hline

\end{tabular}

\label{tab:performance}
\end{table}

\subsubsection{Auditable Data}

We proceed with the feedforward pass through the object classifier network. During this process, we feed the object classifier $M$ with images that have been used to train the network $\mathcal{D}$ and with images the network has not seen before $\mathcal{E}$. For the Model Outcome $y$, we retrieve the object output embedding. 

\subsubsection{MINT Model Training}
We train de MINT Model $T$ for 50 epochs using the Adam optimizer, with batches of 32 samples, each containing an equal number of samples per class. Binary cross-entropy serves as the loss function. The external samples $\mathcal{E}$ from a determinate class are always limiting since they have fewer samples compared to those in training in the object classifier. Therefore, we take as many samples from the used training class $\mathcal{D}$ as the total number of external samples $\mathcal{E}$ from the class to ensure balanced training. This is done in order to achieve a balanced training with the same number of samples from class 0 (External $\mathcal{E}$) and class 1 (training $\mathcal{D}$). There will be as many MINT modules as classes in the database since a separate model is trained for each class. In this case, the metrics used for evaluation have been AUC (Area Under the Curve) and accuracy.

AUC measures the area under the receiver operating characteristic curve (ROC), providing insights into the model's ability to distinguish between positive and negative classes across different thresholds. It's particularly useful when dealing with imbalanced datasets or when the true positive rate and false positive rate are essential considerations as in this case where we have more training samples than test samples that we have to identify. The evaluation of the MINT Module is conducted in a balanced manner \cite{ling2003auc}. Accuracy, on the other hand, assesses the proportion of correctly classified instances over the total number of instances. By utilizing AUC and accuracy as evaluation metrics, we gain a comprehensive understanding of the model's performance in terms of both discrimination ability and overall correctness.

\section{Experiments and Results}\label{sec:experiments}

To better understand the challenges associated to MINT tasks, we first present experiments aimed to study how different training hyperparameters of the object classifier (audited model) affect the performance of the the MINT module.

\begin{figure}[t!]
\centering
\begin{subfigure}[b]{0.49\textwidth}
    \includegraphics[width=\textwidth]{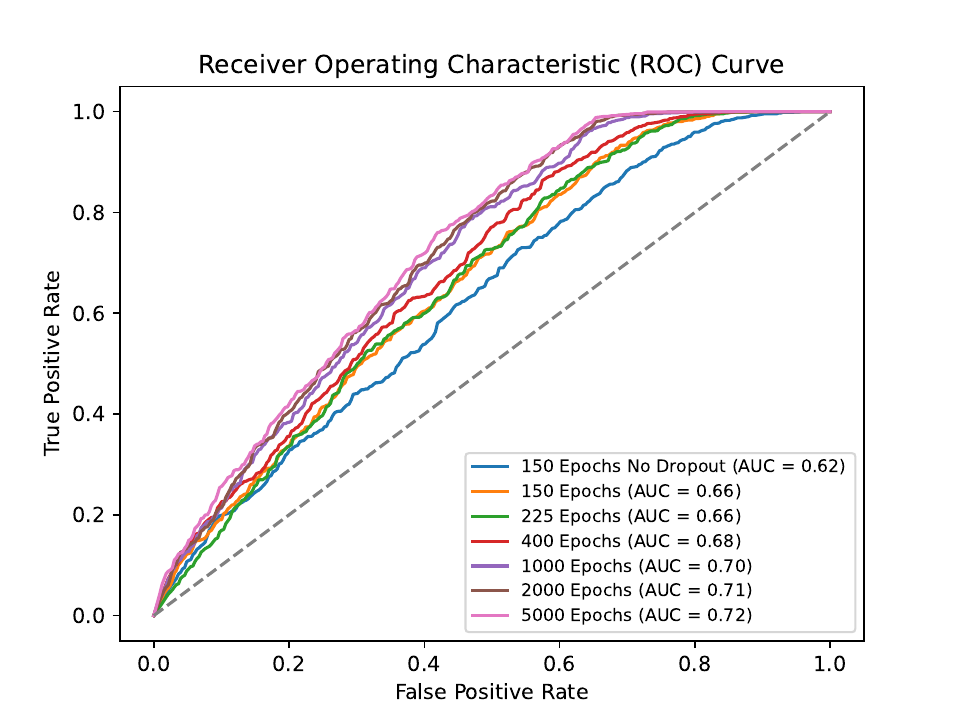}
    \caption{MINT ROC curves evaluated for different epochs over CIFAR-10.}
    \label{fig:rocepochs}
\end{subfigure}
\hfill
\begin{subfigure}[b]{0.49\textwidth}
    \includegraphics[width=\textwidth]{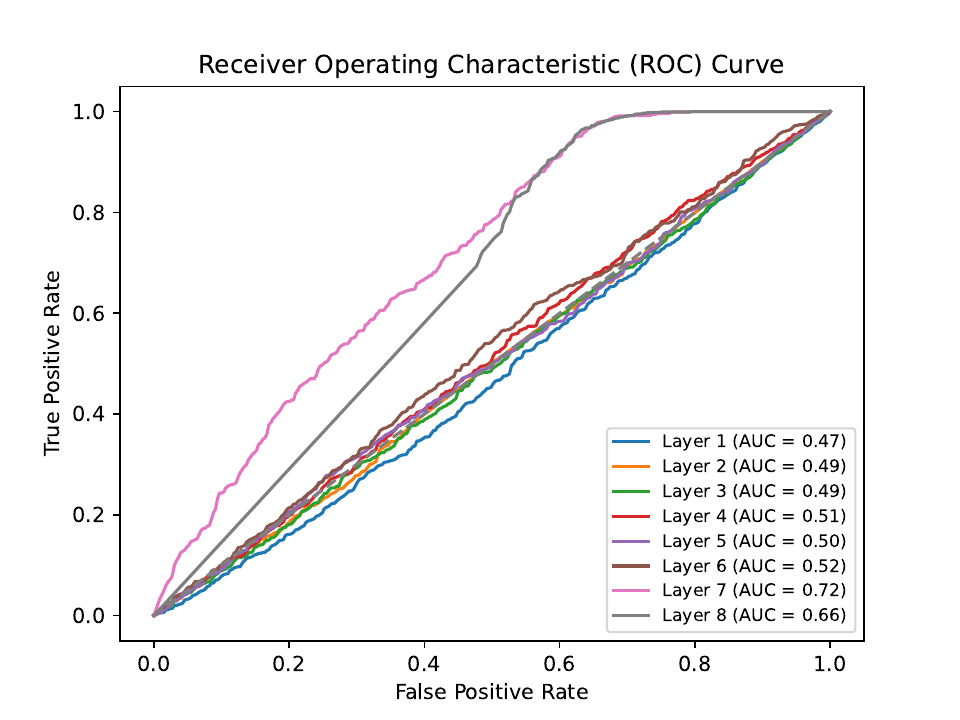}
    \caption{MINT ROC curves evaluated for different layers over CIFAR-10.}
    \label{fig:roclayers}
\end{subfigure}
\caption{MINT ROC curves evaluated over CIFAR-10 for different epochs and layers.}
\label{fig:combined_roc}
\end{figure}

One of the experiments being carried out involves modifying the epochs used to train the audited model. The rationale behind this experiment is to analyze whether the number of times given data is used to adjust the model parameters affects the memorization capacity of the model. The hypothesis is that a larger number of epochs will lead to overfitting of the model and thus more recognizable patterns. This experiment is conducted using the CIFAR10 database, thus altering the number of times the classifier processes the input images without overfitting as can be seen in Table \ref{tab:performance} \cite{yeom2018privacy}. The penultimate layer of this classifier is chosen to process the data through the MINT module, given its consistent superior performance across all potential scenarios, as indicated in Figure \ref{fig:roclayers}.

As shown in Figure \ref{fig:rocepochs}, the performance of the Object Classifier reaches its peak when it has been trained over a total of $5000$ epochs. The experiments show that more epochs increase the training traces, but the AUC tends to stabilize after $5000$ epochs. This indicates that increasing the number of epochs beyond this point does not significantly improve the classifier's performance.

For the rest of the experiments, we will use the audited models trained with 5000 epochs. We conducted an extensive evaluation to assess the performance of different output embeddings from various layers in determining whether samples were utilized for training or testing. Through this evaluation, we aimed to identify the most effective layer and corresponding output embedding for distinguishing between training and testing samples accurately.

\subsection{Impact of the object detector activation layer}

The layer used to extract the activations may affect the performance of the MINT detection. The first layers are traditionally associated with more general image features (such as color, shapes, and textures), while the last layers are traditionally associated with more specific features of the task (e.g., labels).  In Figure \ref{fig:roclayers}, it can be observed that the layer yielding the best results is Layer seven, the penultimate layer, corresponding to embeddings with dimensions 128. The second-best performing layer is the output probability from the classification network with dimension 10, which corresponds to the Layer eight, which is the last layer. These findings suggest that embeddings from last layers, such as those from Layer seven, capture more nuanced features, contributing to better performance in distinguishing between training and testing samples. In contrast, the embeddings from the first layers of a CNN exhibit less discriminative power, as evidenced by their AUC values hovering around 50.

The experiment originally conducted with the CIFAR-10 dataset was replicated, albeit this time utilizing the more complex CIFAR-100 dataset. This replication aims to assess the performance of the model across different layers within its architecture, providing deeper insights into its capabilities across diverse datasets.

\begin{figure}[t!]
\centering

\includegraphics[width=0.55\textwidth]{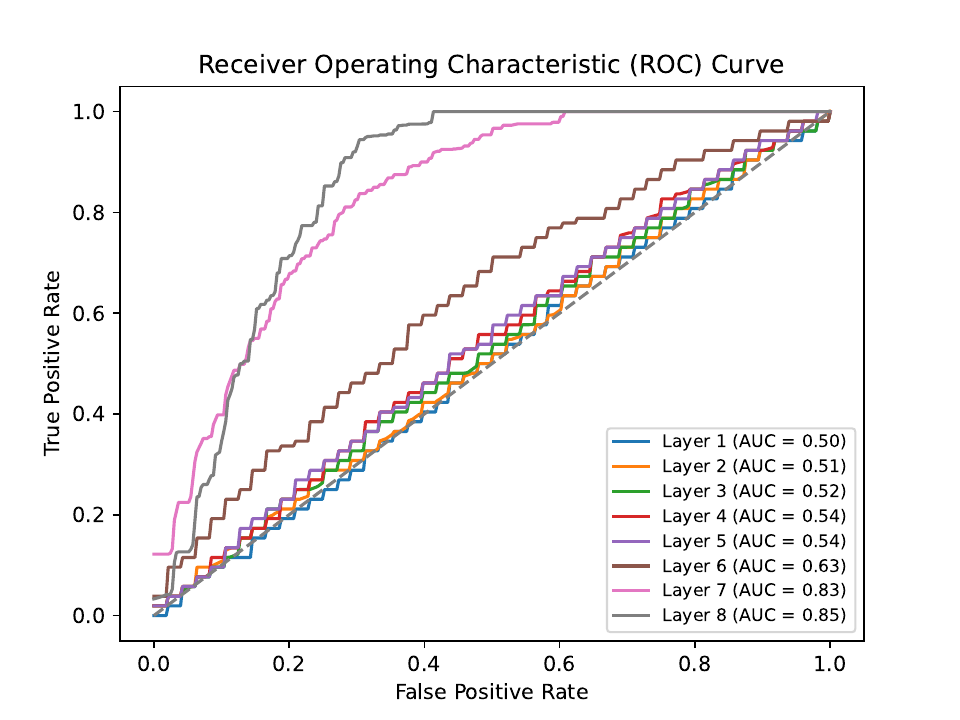}
\caption{MINT ROC curves evaluated for different layers  over CIFAR-100.}
\label{fig:rocclasses100}
\end{figure}

The results depicted in Figure \ref{fig:rocclasses100} provide valuable insights into the performance of the model across its various layers. Interestingly, superior performance is observed in the last two Fully Connected layers, suggesting that these layers are particularly adept at capturing and processing complex patterns within the data. However, the performance of the convolutional layers appears to plateau, with values hovering around 0.5. This observation indicates a certain level of randomness or unpredictability in the outcomes generated by these layers.
\begin{figure}[t!]
\centering
\includegraphics[width=0.55\textwidth]{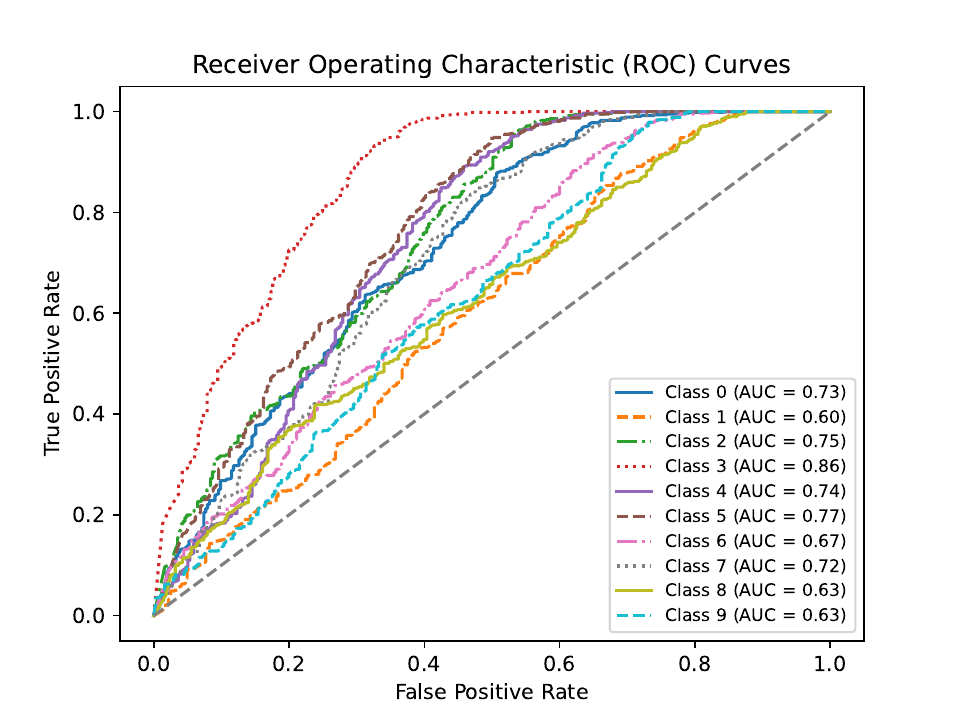}
\caption{MINT ROC curves evaluated for the 10 classes of CIFAR-10.}
\label{fig:rocclasses}
\end{figure}

\subsection{Impact of the object class}

Within our experimental framework, one crucial investigation revolves around the assessment of diverse classes. To achieve this, we  selected the object classifier exhibiting best performance (as can be seen in Figure \ref{fig:rocepochs},the model trained with 5000 epochs) in tandem with the embedding layer configured with a dimensionality of 128, corresponding to the seventh layer, the second-to-last layer. This particular embedding layer consistently outperforms others across multiple evaluation scenarios. By scrutinizing the performance of each class under these settings, we aim to gain comprehensive insights into the efficacy of our MINT system across different categories.

The next experiment is aimed at evaluating whether specific characteristics of the image classes may affect MINT performance. Thus, we analyzed MINT performance for each of the classes in CIFAR-10. As observed in Figure \ref{fig:rocclasses}, the best results were obtained for Classes 2 and Class 3, corresponding to the `bird' and `cat' labels, respectively. The poorest results are observed for Classes 1, 8, and 9, corresponding to the `airplane,' `ship,' and `truck' labels, respectively. Note that all models were trained using the same number of images per class. The experiments suggest that certain characteristics of the classes can affect the accuracy of the MINT model.
 \begin{table*}[h!]
    \caption{Performance of the different MIAs vs our MINT model on CIFAR-10, CIFAR-100 and GTSRB.}
    \centering
   \setlength{\tabcolsep}{2,5pt}

\begin{tabular}{lccllccl}

\hline
\multirow{2}{*}{\begin{tabular}[c]{@{}l@{}}MIAs Methods \end{tabular}} & \multicolumn{2}{c}{AUC}         &                &  & \multicolumn{3}{c}{Balanced Accuracy}            \\ \cline{2-4} \cline{6-8} 
                                                                        & CIFAR-10       & CIFAR-100      & GTSRB          &  & CIFAR-10       & CIFAR-100      & GTSRB          \\ \hline
\multicolumn{1}{l|}{Salem et al. \cite{salem2018ml}}                                       & 0.628          & 0.612          & 0.755          &  & 0.610          & 0.577          & 0.677          \\
\multicolumn{1}{l|}{Yeom at al. \cite{yeom2018privacy}}                                        & 0.646           & 0.804          & 0.818          &  & 0.647          & 0.772          & 0.797          \\
\multicolumn{1}{l|}{Song et al. \cite{song2021systematic}}                                        & 0.644           & 0.804          & 0.820          &  & \textbf{0.650}          & \textbf{0.773}         & 0.681          \\
\multicolumn{1}{l|}{Ye et al. \cite{ye2022enhanced}}                                          & 0.632          & 0.605          & 0.618          &  & 0.527          & 0.578          & 0.606          \\
\multicolumn{1}{l|}{Watson et al. \cite{watson2021importance}}                                      & 0.677          & 0.778          & 0.822          &  & 0.631          & 0.727          & 0.798          \\  \hline
\multicolumn{1}{l|}{MINT [Ours]}                                               & \textbf{0.728} & \textbf{0.826} & \textbf{0.853} &  & 0.649 & 0.745 & \textbf{0.819} \\ \hline
\end{tabular}

\label{tab:auctable}
\end{table*}

\subsection{Impact of the audited model architecture} \label{architeturevariation}

The architecture of the audited model can impact in the performance of the MINT model. An experiment has been conducted in which the classification network was modified to illustrate the performance of MINT models for some of the most popular architectures used in the object classification literature. Therefore, various architectures such as ResNet50\cite{he2016deep}, ResNet100\cite{he2016deep}, and EfficientNetB0 \cite{tan2019efficientnet} have been used. The experimental protocol applied to these architectures has been the same: all have been trained during $1000$ epochs, and the embedding layer (i.e., the last before the output layer) was consistently chosen to maintain a common criterion across all architectures. The experiment was conducted on CIFAR-10.

\begin{table}[h]
\centering
\begin{tabular}{lc}
\hline
                                & \multicolumn{1}{l}{ AUC MINT Module} \\ \hline
\multicolumn{1}{l|}{ResNet50 \cite{he2016deep}}                                          & 0.70                                        \\
\multicolumn{1}{l|}{ResNet100 \cite{he2016deep}}                                         & 0.68                                        \\
\multicolumn{1}{l|}{EfficientNetB0 \cite{tan2019efficientnet}}                                          & 0.77                                        \\
\end{tabular}
\caption{Performance of the architectures for the object classification task and for the MINT module over CIFAR10.}
\label{tab:otherarchitectures}
\end{table}

The results shown in the Table \ref{tab:otherarchitectures} indicate that the MINT approach generalizes for a broad range of architectures. It is understandable that the MINT module exhibits worse performance than previous experiments, given that both the Object classification models and the MINT models were not optimized in search of the best performance.

\subsection{Comparison with MIAs}

We conducted experiments using the databases specified in \ref{sec:framework} (CIFAR-10, CIFAR-100, and GTSRB). Given the novelty of the MINT approach, there are currently no existing works for direct comparison. However, there is a wealth of literature on MIAs, which provides relevant benchmark. Despite the differences in environmental conditions, this field serves as a closely related line of comparison to our MINT approach. We present the results of our study alongside those of MIAs in Table \ref{tab:auctable}. There exists a work on MIAs, proposed by Liu et al. \cite{liu2022membership}, that achieves slightly better performance than those listed in the Table \ref{tab:auctable} across most databases. However, it is not included in the table because it cannot be directly compared to other methods. 
 
This is because a meticulous study of losses and weight variation throughout the entire training process is necessary, as well as utilizing these losses for more precise evaluation. This further implies a deeper understanding of the model being audited. This requirement, while pivotal for the efficacy of Liu's method, presents a challenge as it diverges from the core focus of our own approach where the auditor has partial access to the audited model.

\section{Conclusions} \label{sec:conclusion}

We have designed, implemented, and comprehensive evaluated a Membership Inference Test specifically tailored for machine learning models \cite{herrera23trust} operating within the object classification domain. Our primary aim has been to achieve precise identification of whether a specific sample has been incorporated into the model's training data or not. In pursuit of this objective, we have introduced various experiments to evaluate the sample detection within the dataset. These experiments encompass factors such as the number of epochs employed during the training phase of the audited model, the size of the image database utilized, the resolution of the images, and the strategic selection of model layers. After conducting our study, we achieved a detection rate of approximately 80\% classification accuracy. Furthermore, we identified key factors that facilitate sample recognition. Based on our findings, the results suggest that higher number of classes facilitated the MINT detection. This recommendation stems from the observation that higher image complexity and dataset size tend to enhance the discriminative features available for detection, thereby potentially boosting overall performance.

Our future work in this line will: 1) try to improve the MINT detection accuracy with improved learning architectures, 2) try to exploit general AI models to help in that regard in specific AI/ML domains of our interest such as biometrics \cite{ivan24gpt}, 3) introduce multimodal methods \cite{pena23human} to improve the proposed auditing processes, and 4) investigate how certain human feedback \cite{pena24loop} can help to curate and improve the proposed methods.

\section{Acknowledgment} 

This study has been supported by projects BBforTAI (PID2021-127641OB-I00 MICINN/FEDER) and Cátedra ENIA UAM-VERIDAS en IA Responsable (NextGenerationEU PRTR TSI-100927-2023-2). The work G. Mancera is supported by FPI-PRE2022-104499 MICINN/FEDER. The work of D. deAlcala is supported by a FPU Fellowship (FPU21/05785) from the Spanish MIU. A. Morales is supported by the Madrid Government (Comunidad de Madrid-Spain) under the Multiannual Agreement with Universidad Autónoma de Madrid in the line of Excellence for the University Teaching Staff in the context of the V PRICIT (Regional Programme of Research and Technological Innovation). The work has been conducted within the ELLIS Unit Madrid.

\bibliography{aaai25}

\end{document}